\documentclass{article}

\usepackage{arxiv}

\usepackage[utf8]{inputenc} 
\usepackage[T1]{fontenc}    
\usepackage{hyperref}       
\usepackage{url}            
\usepackage{booktabs}       
\usepackage{amsfonts}       
\usepackage{nicefrac}       
\usepackage{microtype}      
\usepackage{lipsum}

\usepackage{pgfplots}  
\pgfplotsset{width=\linewidth, height=15.5em}
\usepackage{lineno,hyperref}
\usepackage{diagbox}
\usepackage{booktabs}
\usepackage{multirow}
\usepackage{boldline}
\usepackage{rotating}
\usepackage{lscape} 
\usepackage[sort&compress,square,comma,numbers]{natbib}

\title{PREDICT: Persian Reverse Dictionary}

\author{
  Arman Malekzadeh \\
  Department of Mathematics and Computer Science\\
  Amirkabir University of Technology\\
  Tehran, Iran \\
  \texttt{arman.malekzade@aut.ac.ir} \\
   \And
 Amin Gheibi
 \thanks{Corresponding Author} \\
  Department of Mathematics and Computer Science\\
  Amirkabir University of Technology\\
  Tehran, Iran \\
  \texttt{amin.gheibi@aut.ac.ir} \\
   \And
 Ali Mohades \\
  Department of Mathematics and Computer Science\\
  Amirkabir University of Technology\\
  Tehran, Iran \\
  \texttt{mohades@aut.ac.ir} \\
}

\begin{document}
\maketitle

\begin{abstract}
Finding the appropriate words to convey concepts (i.e., lexical access) is essential for effective communication. Reverse dictionaries fulfill this need by helping individuals to find the word(s) which could relate to a specific concept or idea. To the best of our knowledge, this resource has not been available for the Persian language. In this paper, we compare four different architectures for implementing a Persian reverse dictionary (PREDICT). 
 We evaluate our models using \textit{(phrase,word)} tuples extracted from the only Persian dictionaries available online, namely Amid, Moein, and Dehkhoda where the \textit{phrase} describes the \textit{word}. Given the \textit{phrase}, a model suggests the most relevant word(s) in terms of the ability to convey the concept.
The model is considered to perform well if the correct \textit{word} is one of its top suggestions.
  Our experiments show that a model consisting of Long Short-Term Memory (LSTM) units enhanced by an additive attention mechanism is enough to produce suggestions comparable to (or in some cases better than) the \textit{word} in the original dictionary. The study also reveals that the model sometimes produces the synonyms of the \textit{word} as its output which led us to introduce a new metric for the evaluation of reverse dictionaries called \textit{Synonym Accuracy} accounting for the percentage of times the event of producing the \textit{word} or a synonym of it occurs. The assessment of the best model using this new metric also indicates that at least 62\% of the times, it produces an accurate result within the top 100 suggestions.
\end{abstract}


\section{Introduction}
A reverse dictionary is a tool which provides a user with the closest word(s) to a concept, given a descriptive phrase corresponding to it. For instance, the tool might submit \textit{venus} as the output word, given \textit{the nearest planet to earth} as the input phrase.

Language producers (i.e., writers) could use this special type of dictionary to express a set of ideas using a single word \citep{calvo2016integrated}.
Also, it is believed that reverse dictionaries can serve as a means of the treatment of \textit{word selection anomia aphasia} \citep{rohrer2008word} which is a disorder causing patients to be unable to name an object although having the ability to describe and identify it \citep{thorat2016implementing}.
In addition, bilingual reverse dictionaries \citep{lam2013creating} have been used to expand our knowledge of the resource-poor languages or protect the endangered ones. Given a phrase in a resource-rich language (e.g. English), they provide us with a term in a resource-poor one (e.g. Vietnamese) or vice versa. 
Moreover, questions answering and information retrieval tasks (i.e., associating medical concepts with input descriptions in text format) have been tackled by reverse dictionary models \citep{kartsaklis2018mapping}.
Also, previous research shows that reverse dictionaries are capable of solving crossword puzzles by understanding phrases  \citep{hill2016learning}. 

Researchers have developed digital reverse dictionaries for many languages including English \citep{zock2004word}, Japanese \citep{bilac2004dictionary}, Turkish \citep{el2004use}, and French \citep{dutoit2002lexical}. Yet, we found no attempt on developing a similar tool for the Persian language which might be due to the lack of appropriate and large enough datasets containing words and their corresponding definitions. We found only three accredited \textit{forward} dictionaries (Amid \citep{amid}, Moein \citep{moin}, and Dehkhoda \citep{dehkhoda}) available online and in a machine-readable format, and chose them to be used as our primary sources of information on the words of the Persian language. According to the definition of low-resource languages given by \citet{duong2017natural}, Persian can be considered a low-resource language for developing reverse dictionaries since there is no algorithm using currently available data to automatically develop a Persian reverse dictionary with adequate performance.


In this paper, we study different deep neural networks which map a descriptive phrase to a word in order to simulate the functionality of a reverse dictionary. In order to train our models, \textit{(phrase, word)} tuples are needed. We use three Persian language resources to extract these tuples: the top three most popular Persian dictionaries \citep{amid,dehkhoda,moin}, Farsi Wikipedia, and Farsnet \citep{shamsfard2010semi}.

Due to the fact that a word may have multiple senses, the extracted data may contain several phrases describing it. The models should map all of these different (and sometimes irrelevant) phrases to the same word. Also, they are expected to learn the mapping regardless of the preffered grammatical style and word choice of the authors of the phrases who have lived in different times.


Our experiments reveal that enhancing the RNN model proposed by \citet{hill2016learning} with an attention mechanism \citep{bahdanau2014neural} is enough to produce words comparable to (and in some cases better than) the ones in the dictionary tuples.

The rest of this paper is organized as follows. In section \ref{sec:rel-work}, we summarize the approaches used to design reverse dictionaries. Section \ref{sec:proposed-approach} explains our proposed approach. Section \ref{sec:experiments} provides information about out experiments, and section \ref{sec:conclusion} concludes them in addition to supplying the reader with the ideas we believe could be useful for future work.

\section{Research Objectives \label{sec:research-objectives}}
This paper aims to investigate the capabilities of artificial neural networks in learning to map descriptions to words in a vector space using dictionary data in Persian. The focus will be on the use of LSTM \cite{hochreiter1997long} neural networks and the attention mechanism \cite{bahdanau2014neural} for tackling this task. In this study, we are particularly interested in evaluating the performance of our models using both purely computational metrics and other methods relying on the opinions of the experts in linguistics who have seen the output words and scored them. 

\section{Related Work \label{sec:rel-work}}
Many of the reverse dictionary models use a knowledge graph in which nodes represent words and edges demonstrate the relations between them \citep{ferret2006enhancing}. In some of these models, the similarity between two words is determined by the length of a shortest path between their corresponding nodes. The relatedness of a phrase and a word $w$ could then be calculated by different techniques, i.e., considering a node for the phrase \citep{dutoit2002lexical} or averaging the similarity of the words belonging to the phrase with $w$ \citep{thorat2016implementing}. Another model by \citet{mendez2013reverse} represents words and phrases as vectors based on semantic analysis using a Wordnet \citep{miller1995wordnet} and measures their resemblance with metrics such as cosine similarity.

Another group of the researchers have tried to leverage electronically-stored corpora. The simplest models of this type directly compare the input phrase with dictionary definitions \citep{el2004use,shaw2011building} while more advanced ones first calculate a vector to represent both the phrase and the definitions \citep{bilac2004dictionary}. \citet{zock2004word} have also proposed a method which computes the similarity of the phrase and a word by considering the number of paragraphs in Wikipedia pages which contain both the word and the terms existing in the phrase.

The new generation of reverse dictionaries employs word embeddings along with artificial neural networks to enhance the process of mapping the input phrase to a word. \citet{hill2016learning} introduced the first model of this kind by employing an LSTM \citep{hochreiter1997long} which learns to map the constituent words of a phrase describing another word to a vector close to its embedding. \citet{morinaga2018improvement} improved this model by adding the capability of inferring the category of the input using a CNN \citep{kim2014convolutional} and pruning the irrelevant words accordingly. \citet{pilehvar2019importance} also achieved better performance using a unique embedding for each sense of a word. These embeddings were used by \citet{hedderich2019using} along with an attention layer which decides which embedding should be used for each word in the input phrase. \citet{zheng2020multi} changed the simple LSTM used by \citet{hill2016learning} to a bidirectional one (BiLSTM). By using an attention layer on top of the BiLSTM, their model is able to estimate the amount each word of the input phrase contributes to its whole meaning. Using single-layer perceptrons, it can also predict the part-of-speech tag and the written form of the correct output word.
Recently, \citet{morinaga2020improvement} performed experiments to reach a solution by which the capacity of a neural reverse dictionary (the kind in which a neural network is employed) can be improved. Supported by these experiments, they claimed that adjusting the output of an LSTM layer by adding bypass structures called Cascade Forward Neural Networks (CFNN) is more effective than simply adding more neurons to the LSTM layer. Another recent enhancement was done by \citet{qi-etal-2020-wantwords} who used BERT \cite{devlin-etal-2019-bert} to represent sentences in a vector space. Since the tool they designed can take a description in English, convert it to a phrase in Chinese using the Baidu Translate API \footnote{\href{https://fanyi-api.baidu.com/}{https://fanyi-api.baidu.com/}}, and produce the target word in Chinese, it is called a \textit{bilingual} reverse dictionary. To bridge the gap between the languages in this type of dictionaries, another method is to use cross-lingual word embeddings \cite{chen-etal-2019-learning-represent}. 

\section{Proposed approach \label{sec:proposed-approach}}
The merits of deep neural networks in mapping a phrase to a word capable of conveying the concepts within it have been shown by the recent studies \citep{hill2016learning,kim2014convolutional,hedderich2019using,zheng2020multi}. Therefore, we find them suitable for implementing a Persian reverse dictionary (PREDICT).
\subsection{Model Architecture}
As baseline reverse dictionary models, we use the BOW and RNN models designed by \citet{hill2016learning}. We also propose two new models for the reverse dictionary implementation.

All of the models aim to map a phrase, i.e., a sequence of words $p=(w_1, w_2, \dots, w_k)$ to the word $w$ it describes. Formally, each model accepts a sequence of vectors $v_1, v_2, \dots, v_k$ where $v_i \in \mathbb{R}^{d}$ is the embedding of the word $w_i$ in the phrase $p$, and outputs another vector $v_o \in \mathbb{R}^d$. We wish for $v_o$ to be close to the embedding of the word $w$ which we denote by $v_w \in \mathbb{R}^{d}$. To describe the models, we first need to introduce the additive attention mechanism \citep{bahdanau2014neural}.
\subsubsection{Additive Attention}
Suppose we have a sequence of vectors $u=(u_1, u_2, \dots, u_N)$ each belonging to $\mathbb{R}^{d}$, and another vector $q\in\mathbb{R}^{d}$. If we refer to $u_i$'s the values and to $q$ as a query, the additive attention can be seen as a function of the query and the values:
\begin{equation}
attention(q,u)=\sum_{i=1}^{N} \alpha_i u_i
\end{equation}
where $\alpha_i \in \mathbb{R}$ is the attention assigned to the $i$-th value. To obtain the value of $\alpha_i$, first an attention score $e_i \in \mathbb{R}$ is calculated:
\begin{equation}
e_i = \tanh(u_i+q) \quad t=1,2,\dots,N
\end{equation}
Then, a vector $e\in\mathbb{R}^{N}$ containing all of the scores is formed to calculate $\alpha \in\mathbb{R}^{N}$:
\begin{equation}
e=(e_1,e_2,\dots,e_N)
\end{equation}
\begin{equation}
\alpha=softmax(e)=(\alpha_1,\alpha_2,\dots,\alpha_N)
\end{equation}
\subsubsection{LSTM with Attention}
In this model, the input vectors pass through an LSTM \citep{hochreiter1997long} one at a time. We denote the hidden state of the LSTM at time $t$ by $h_t\in\mathbb{R}^{d\times 1}$. The model projects the hidden states to the same space linearly to calculate a value $p_t$ for each moment of time:
\begin{equation}
p_t = W h_t + b \quad t=1,2,\dots,k
\label{eq:lstmatt-first-proj}
\end{equation}
Additionally, it calculates another projected value $p'_k$ for the last hidden state:
\begin{equation}
p'_k = W' h_k + b'
\label{eq:lstmatt-second-proj}
\end{equation}
In equations \ref{eq:lstmatt-first-proj} and \ref{eq:lstmatt-second-proj}, both $W$ and $W'$ are projection matrices belonging to $\mathbb{R}^{d \times d}$ while $b$ and $b'$ are bias vectors in $\mathbb{R}^{d\times 1}$.
Since the words in the input phrase contribute to its meaning differently, we apply an additive attention mechanism on the projections to obtain another vector $a\in\mathbb{R}^{d\times1}$:
\begin{equation}
a=attention(p'_k, (p_1,p_2,\dots,p_k))
\label{eq:attention-on-projections}
\end{equation}
After the calculation of $a$, the model projects it into the same space and applies the hyperbolic tangent function to each element of it to obtain the value of the output $v_o$:
\begin{equation}
v_o = \tanh(W'' a + b'')
\label{eq:lstm-output}
\end{equation}
where  $W''\in\mathbb{R}^{d\times d}$ is a projection matrix, and $b''$ is a bias vector.
\subsubsection{BiLSTM with Attention}
This model uses a Bidirectional LSTM (BiLSTM) \citep{schuster1997bidirectional} to process the input vectors. The BiLSTM produces two sequences of hidden units $h_1,h_2,\dots,h_k$ and $h'_1,h'_2,\dots,h'_k$ with respect to its forward and backward processing. The corresponding elements of the two sequences are then concatenated together to form a unified sequence:
\begin{equation}
c_t = concat(h_t, h'_t) \in \mathbb{R}^{2d\times1} \quad t=1,2,\dots,k
\end{equation}
The model then projects each of the $c_t$'s linearly to the initial space $\mathbb{R}^{d}$:
\begin{equation}
p_t = W c_t+b \quad t=1,2,\dots,k
\label{eq:bilstmatt-first-proj}
\end{equation}
It also projects the last concatenated value $c_k$ to the initial space in the same way separately:
\begin{equation}
p'_k = W' c_k+b'
\label{eq:bilstmatt-second-proj}
\end{equation}
In equations \ref{eq:bilstmatt-first-proj} and \ref{eq:bilstmatt-second-proj}, $W$ and $W'$ are projection matrices and belong to $\mathbb{R}^{d\times 2d}$ while $b$ and $b'$ are bias vectors in $\mathbb{R}^{d\times1}$.

The additive attention mechanism is then applied to the projected vectors as in equation \ref{eq:attention-on-projections} and the output is calculated as in equation \ref{eq:lstm-output}.
\subsection{Loss Function}
Presuming that the model has accepted a phrase $p$ describing a word $w$ and produced a vector $v_o$, we define the cosine loss of the model on the tuple $(p,w)$ to be the cosine distance between $v_o$ and $v_w$ where $v_w$ is the embedding of $w$. Consequently, we  set the cosine loss of the model on a set of tuples to be the average of its loss on the individual tuples. 
\subsection{Evaluation Metrics \label{sec:eval-metrics}}
Assume that $v_1,v_2,\dots,v_k$ is a sequence of word vectors of a phrase $p$ describing a word $w$ and $v_o$ is the output of the model. We use various metrics to evaluate the quality of the output vector ($v_o$).

Suppose that $V$ is the list of the words recognized by the model. We sort $V$ based on the cosine similarities of the embeddings of its elements to $v_o$ in the descending order, and refer to the result as the output ranking. We also refer to the $i$-th word in the ranking as the model's $i$-th suggestion ($s_i$), and to the word described by the phrase as the original word.

For a list of phrases and their corresponding original words, the model produces a list of rankings each consisting of a list of suggestions. We use three metrics for the evaluation of these suggestions.
\begin{enumerate}
	\item \textbf{Accuracy} 
	
	Inspired by \citet{hill2016learning}, we consider the proportion of phrases for which the original word is in the first 10/100 suggestions (\textit{accuracy@10/100}) as a metric to evaluate the model. Note that if two words are synonyms and one of them conveys the meaning of a descriptive phrase, the other one may also have the same capability (e.g., `\textit{football}' and `\textit{soccer}' both can refer to `\textit{a game played using a ball between two teams each consisted of eleven players}').
	However, measuring this metric completely ignores existence of synonyms of the original word in the top suggestions, while a human may accept them as valid words conveying the same concepts.
	
	\item \textbf{Synonym Accuracy}
	
	We store a set of synonyms for each word (see section \ref{sec:secondary-datasets} for more details).
	We define \textit{synonym-accuracy@10/100} to be the proportion of phrases for which the original word or a synonym of it exists in the first 10/100 suggestions. Measuring this metric helps with mitigating the problem of ignorance of synonyms with the accuracy metric.
	
	\item \textbf{Mean Opinion Score \label{item:mos}}
	
	To assess the quality of the suggestions in practice, we also ask a group of 29 Persian linguistics experts to rate the original word and the first three suggestions of the model for each phrase. Each person assigns a score $q\in\mathbb{N}$ from 1 to 5 to each suggestion and the original word. Before the beginning of the rating, we explained the meaning of each possible score to the raters. More specifically, we informed them that the scores from 1 to 5 mean `\textit{completely irrelevant}', `\textit{slightly relevant}', `\textit{surely relevant}', `\textit{acceptable match}' and `\textit{perfect match}' respectively. To clarify our notion of these meanings, we used some examples from the real dictionaries.   
	
	After two raters $r$ and $r'$ have announced their scores for a list of identical items, we use the weighted kappa coefficient \citep{cohen1968weighted} to quantify the agreement between them (denoted by $\kappa(r,r')$). We employ linear weights \citep{cicchetti1971new} for the calculation of the coefficient. To measure the \textit{agreement} between each rater $r$ and the others $r'_1,r'_2,\dots,r'_m$, we compute the average of the agreement of $r$ with each of them:
	\begin{equation}
	agreement(r, (r'_1,r'_2,\dots,r'_m)) = \frac{\sum_{i=1}^{m} \kappa(r,r'_i)}{m}
	\label{eq:agreement-score}
	\end{equation}
	\citet{landis1977measurement} have claimed that a value from 0.41 to 0.6 shows a moderate agreement between two raters. Considering this range, we define a \textit{valid rater} to be a person with an agreement of at least 0.41 with the others.
	
	For each list of scored items, we first identify the valid raters. Considering only the scores announced by these raters, we remain with 27 valid raters on average. We calculate the mean score given by them to each item. This value is called the mean opinion score (MOS) given to the item.
\end{enumerate}

\section{Experiments \label{sec:experiments}}
\subsection{Data}
We experiment with multiple Persian datasets to train and evaluate our models.
We have six primary datasets and three secondary datasets (obtained from the primary datasets).
\subsubsection{Primary Datasets}
\begin{itemize}
	\item \textbf{Vajehyab} \footnote{\url{http://vajehyab.com}} is a website containing the lexical entries of many Persian dictionaries including Amid, Dehkhoda, and Moin in HTML format. Using a web crawler, we stored 666,941 pages of the website.
	Each page corresponds to a lexical entry providing information about a word including its senses, pronunciation, and part-of-speech tags. We will refer to this word as the main word of the page in the rest of this paper.
	\item \textbf{Persian Wikipedia Articles} are available in XML format on Wikimedia \footnote{\url{https://dumps.wikimedia.org/backup-index.html}}. 
	We downloaded a dump of the Persian Wikipedia containing 609,850 articles written until March 16, 2020. For each article, the title, the text of each section, and the interlinks are available.
	\item \textbf{Farsnet} is a Persian Wordnet developed by 
	\citet{shamsfard2010semi}. The third version of the Farsnet containing information about 100,000 words is available online and accessible via its API \footnote{\url{http://farsnet.nlp.sbu.ac.ir}}. The set of phrases describing each word is also included in the information.
	\item \textbf{Uppsala Persian Corpus}
	is a modified version of the Bijankhan Corpus \citep{bijankhan2004role} published by
	\citet{seraji2015morphosyntactic} consisting of 2,704,028 tokens along with their part-of-speech tags.
	\item \textbf{Hamshahri Corpus}
	is based on the news articles published in the online version of Hamshahri newspaper. Totally, there are 166,774 documents made of 417,339 unique terms in the corpus. Each document is filled with information about an article including an identifier, a publication date, a category and some text \citep{aleahmad2009hamshahri}.
	\item \textbf{Comprehensive Dictionary of Synonyms and Antonyms of the Persian Words} contains 15,000 lexical entries in 27400 semantic categories. Each entry provides the set of synonyms and antonyms of a particular word. We use an electronic version of the dictionary available on Peykaregan website \footnote{\url{https://www.peykaregan.ir/node/324?rid=252}}.
	
\end{itemize}
\subsubsection{Secondary Datasets \label{sec:secondary-datasets}}
\paragraph{Persian Words Ranking \label{sec:words-ranking}}
We design an experiment to approximate the frequency of each Persian word. The frequency of a word in a large Persian corpus can be considered as a fair estimate of its true frequency in the Persian language. To create such a corpus, we combine the following parts of our primary datasets:
\begin{itemize}
	\item The lexical entry found in each HTML page of the Vajehyab website (after removing the HTML tags)
	\item The full text of Wikipedia pages
	\item The tokens of the Uppsala Persian Corpus
	\item The text of the articles in the Hamshahri Corpus
\end{itemize}

\begin{figure}[h]
	\centering
	\caption{An example of the structure of a JSON object containing a  word along with its corresponding phrases and the source they were extracted from.}
	\includegraphics[width=0.8\linewidth]{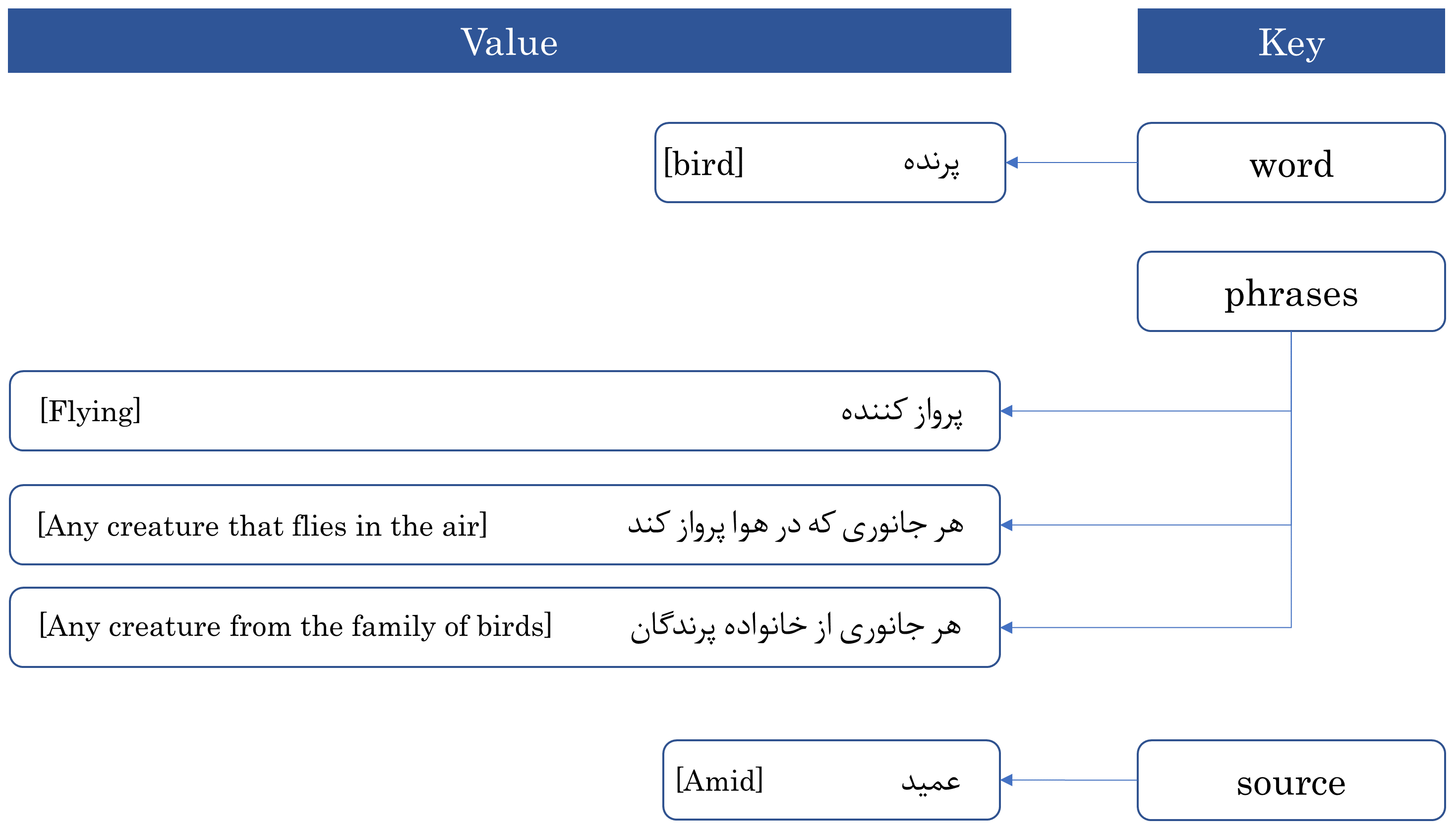}
	\label{fig:json-obj-ex}
\end{figure}

Many letters of the Persian alphabet are derived from the Arabic one. Therefore, it is common to see the Arabic form of a letter in Persian corpora. For instance, the Persian letter ``\includegraphics{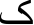}'' (kāf) is written as ``\includegraphics{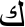}'' in Arabic and has the same name and pronounciation. Another example would be the Arabic letter ``\includegraphics{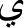}'' (yā) for which the Persian equivalent is the letter ``\includegraphics{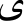}'' (ye).
After creating the corpus, we normalize it by replacing the Arabic characters with their Persian equivalent (if available) and removing the non-Persian characters.
Then, we store the list of words in the normalized corpus sorted by their frequency. We will use this list as a ranking of the Persian words.
\paragraph{Extracted Phrases}
We extract a word and a list of phrases describing each sense of it from each page of the Vajehyab website. Then, we save the list of the descriptive phrases together with the word they correspond to in a JSON object having three keys: word, phrases and source where the source is the dictionary that the lexical entry belongs to (Amid, Dehkhoda, or Moin). The result of this step is a list of 834,750 such objects. We noticed that some of the pages have contained the same content, and some were empty. Therefore, we removed the corrupted (or useless) objects resulted from these pages, and remained with 567990 objects. We will refer to these objects as the Vajehyab objects in the rest of this paper.
Figure \ref{fig:json-obj-ex} demonstrates a sample of the gathered objects.

Also, we consider the sentences of the first section of each Wikipedia page as descriptors for the word appearing in the title of the page.
 Using this process, we obtain a list of 606,197 JSON objects with the same structure as the ones extracted before (each having \textit{Wikipedia} as their \textit{source}). We then consider the list of all words described in the dictionaries, and remove the objects in which another word is described. The result is a total of 23728 objects.

Additionally, we tried to collect the information of the top 30000 frequent words based on our previously made ranking via the API of the Farsnet (Persian Wordnet). However, the resource only had the information of 21123 words. We stored the descriptive phrases of each word along with itself in a JSON object as before and made another list of JSON objects (with \textit{Farsnet} considered as the \textit{source} for all of them). 
\paragraph{The Integral Set of Synonyms}
We extract the set of all synonyms for each word from the comprehensive dictionary.
Also, we expand this set by considering a word $w_1$ as a synonym of another word $w_2$ if there exists a JSON object having $w_2$ as the word and $w_1$ as one of the phrases. We will refer to this modified set as the integral set of synonyms.
\subsubsection{Preprocessing}
We apply several preprocessing steps to the extracted phrases before preparing them for the training and evaluation of the models.

First, we normalize the characters within the extracted phrases and their corresponding words, similar to section \ref{sec:words-ranking}. Then, we replace psuedo-spaces found in the definition of a word with white spaces (e.g., ``\includegraphics{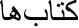}'' gets replaced by ``\includegraphics{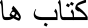}''). We also prevent each word from defining itself by removing it from the phrases which describe it.

Moreover, we remove each JSON object containing the descriptive phrases for a stopword, using a list of 185 stopwords collected manually. In addition, we omit the stopwords from the descriptive phrases existing within each object.

We remove the short phrases (having less than three tokens) from each object. We also remove the objects corresponding to a word with less than three characters, since a Persian speaker does not need an intelligent system to recall such a short word.

Afterwards we tokenize the phrases within the objects based on the occurrence of the space character. We then make a list of all unique tokens and sort it by the frequency of them in the descending order. Finally, we normalize the tokens by only considering the top 100,000 elements in the list as the recognized tokens and removing all of the others from the phrases.
\subsubsection{Preparation}
In order to make the data readable for our models, 
we begin by transforming each JSON object into a list of $(p,w)$ tuples where $p$ is a phrase the object contains and $w$ is the word the phrase describes. We then randomly split the tuples resulted from transforming the Vajehyab objects in the ratio of 8:1:1. We call these proportions the training, development, and testing tuples respectively. We extract tuples from the JSON objects derived from Farsi Wikipedia and Farsnet in the same way and add all of them to the training tuples.
\begin{figure}
	\centering
	\caption{The data preprocessing and preparation pipeline}
	\includegraphics[height=300px]{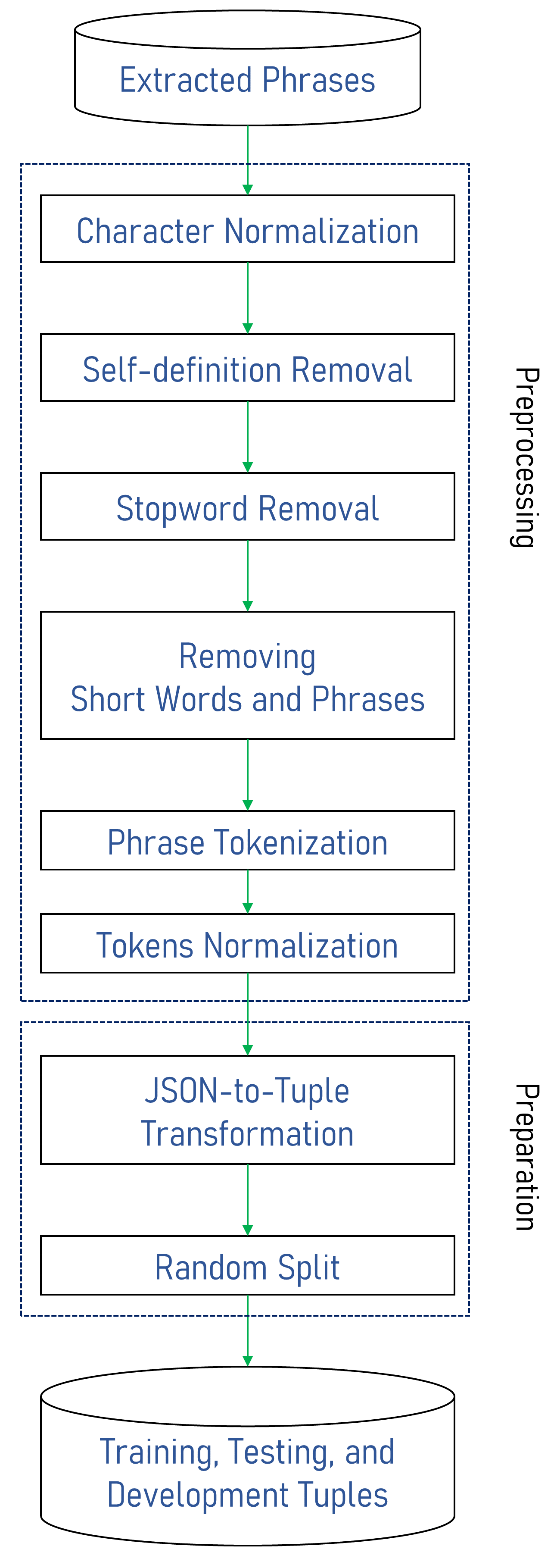}
	\label{fig:data-pipeline}
\end{figure}
Figure \ref{fig:data-pipeline} summarizes all of the steps for the data preprocessing and preparation.

\subsection{Text Coverage Measurement \label{sec:text-coverage-measurement}}
Training a model with a lot of samples may result in better generalization. However, the more training samples we have, the more time each model needs to be trained. To reach a trade-off between the training time and the generalization of the models, we design an experiment to estimate the proportion of Persian text producible by the most frequent words.

We first prune the Persian words ranking so that it does not contain any stopwords or a word without a lexical entry in a dictionary.
We also get the full text of Wikipedia pages and normalize its characters as  in section \ref{sec:words-ranking}.
We remove all of the stopwords from the text and convert each word to its lemma. Then, we choose the top $n$  words remaining in the ranking and measure the percentage of the lemmas they cover\footnote{We consider the frequency of each lemma in this measurement.}.

Figure \ref{fig:wiki-coverage} shows that there is no significant improvement in the coverage as we increase the number of words ($n$) from $20000$ to higher values.

\begin{figure}[ht]
	\caption{Persian Wikipedia Coverage Based on the Word Ranking}
	\resizebox{1.0\columnwidth}{!}{%
		\begin{tikzpicture}
		\begin{axis}[
		ybar,
		ylabel={Lemma Coverage (Percents)},
		xlabel={The Number of Words (Kilos)},
		ymax=100,
		ymin=0,
		x label style={at={(axis description cs:0.5,-0.05)}},
		symbolic x coords={3,5,10,20,30,40,50,60},
		xtick=data,
		nodes near coords,
		nodes near coords align={vertical},
		nodes near coords style={font=\footnotesize}
		]
		\addplot coordinates {(3,69) (5,75) (10,80) (20, 83) (30, 84.1) (40, 84.3) (50, 84.5) (60, 84.5)};
		\end{axis}
		\end{tikzpicture}
	}
	
	\label{fig:wiki-coverage}
\end{figure}

Hence, we will train and evaluate each model only with the $(p,w)$ tuples having one of the top 3000, 5000, 10000, or 20000 words from the ranking as their second element. We call these words the recognized output words of the model. Table \ref{tab:tr-de-te} represents the number of training, development, and testing tuples remaining after applying this restriction on the number of words.

\begin{table}
	\centering
	\begin{tabular}{l|*{4}{c}}
		\toprule
		\diagbox{\# Words}{\# Samples} & Training & Development & Testing \\\hline
		\midrule
		3000 & 101796 & 9352 & 8049  \\\hline
		5000 & 146203 & 14657 & 12210  \\\hline
		10000 & 234650 & 24902 & 20344  \\\hline
		20000 & 337984 & 37378 & 30646  \\\hline
		\bottomrule
	\end{tabular}%
	\caption{The number of tuples remaining considering the most frequent words}
	\label{tab:tr-de-te}
\end{table}%

\subsection{Experimental Setting}
We use the Persian FastText embeddings published by \citet{grave2018learning} to initialize the input vectors for each model, but allow them to be adjusted during the training phase. We initialize all of the other trainable parameters of the models using Glorot uniform method \citep{glorot2010understanding}. We use Adam \citep{kingma2014adam} as our optimization strategy with an initial learning rate of 1. Moreover, we consider training batches of size 16. The number of epochs needed to train the models is determined by the early-stopping technique \citep{prechelt1998early}. Also, for each $(p,w)$ tuple in the training, development, or testing tuples, we compare the output vector ($v_o$) with the FastText embedding of the word $w$. However, this embedding is fixed during the training and evaluation phases. All of the models were implemented using Tensorflow \cite{abadi2016tensorflow}.
\subsection{Stratified Sampling for Evaluation}
We design a procedure to choose $s$ samples from a list of tuples to evaluate each model. Assume that we have trained the model using $t$ tuples each having one of the top $n$ frequent words (as we described in section \ref{sec:text-coverage-measurement}) as their second element. We first split these tuples into $s$ buckets $b_1, b_2, \dots, b_s$ where the $i$-th bucket ($b_i$) consists of the tuples having a second element (word) ranked between $\lceil(i-1)(\frac{t}{s})\rceil$ and $\lfloor(i)(\frac{t}{s})\rfloor$. Then, we randomly choose a tuple from each bucket. The chosen tuples are used for evaluation.

\subsection{Evaluation}
We choose an stratified sample of $500$ tuples to assess the \textit{accuracy} and \textit{synonym accuracy} of each model on the training (seen) and testing (unseen) tuples. Table \ref{tab:acc-syn} presents the cosine loss, \textit{accuracy@10/100} and \textit{synonym accuracy@10/100} of the models based on the number of their recognized output words.

Also, for each dictionary we choose another sample of $40$ testing tuples extracted from the JSON objects having the dictionary as their source. We design a procedure to perform a comparison between the model's suggestions and the original word existing in the dictionary. We consider the MOS value (described in section \ref{sec:eval-metrics}) assigned to the $i$-th suggestion of the model for each chosen tuple. We report the average of these values as the quality of the $i$-th suggestions given by the model ($q_i$). We similarly define the quality of the original words existing in the tuples ($q_t$) as the average of their corresponding MOS values. Table \ref{tab:mos-eval} compares the quality of the suggestions of the models with the quality of original words in the dictionaries.

\section{Discussion \label{sec:discussion}}
\subsection{Representation Learning}
The problem of representing words and the sequences consisted of them has been approached by various methods including learning by the help of large corpora indicating the scenarios in which a word can be used \cite{Mikolov2013EfficientEO}. Our models can also be thought of as tools of approaching the same problem.
Since each $v_i$ (input vector) corresponds to a unique word, the set of all unique $v_i$'s can be considered as learned word representations after the training of a model is finished.
Also, the output vector $v_o$ can be seen as a representation for the sequence of words given as input.
We conjecture that our models recognizing a comprehensive list of words and their senses gathered by the experts in linguistics in the course of many years can perform better than the previous models on the task of representing rare or obsolete word senses since they remain infrequent or absent even in large corpora that are available in digital formats. Despite this possible advantage, all of our models suffer from a way of representing out-of-vocabulary (OOV) words. This problem has been mitigated by considering additional vectors for the smaller parts of each word (called n-grams) so that each OOV word consisting of the seen n-grams can still be modeled \cite{bojanowski-etal-2017-enriching,peters-etal-2018-deep,devlin-etal-2019-bert}. However, these n-grams for which the mentioned models capture linguistic features might not even have a meaning in the language. For example, consider the word \textit{playing} for which the n-grams including \textit{ing} and \textit{pla} have been represented by vectors. While \textit{ing} is a real postfix used in many words and delivers specific information about them, \textit{pla} in \textit{playing} does not have anything to do with the complete word itself. As discussed in \ref{sec:text-coverage-measurement}, instead of carrying the overhead of modeling these meaningless n-grams causing a model to have a larger volume, we consider only the set of real words which are used more frequently and cover most of the available large corpora for a language as the vocabulary of our models.
\subsection{Reverse Dictionary Evaluation Metrics}
Most of the previous reverse dictionaries have been evaluated using information retrieval metrics such as precision \cite{reyes2019designing}, recall \cite{shaw2011building}, and accuracy \cite{hill2016learning}. Unfortunately, these concepts do not specify how far the outputs of a reverse dictionary are from being perfectly correct. For instance, all of them consider \textit{Kangaroo} as a wrong suggestion for the phrase \textit{a tall animal usually found in Africa}. However, there should be a method which captures the fact that \textit{Kangaroo} is a tall animal, and is a better suggestion than \textit{chicken} in this case. We tried to compensate for this lack by considering the synonyms of the perfect output for each query as another correct output. However, there still remains the lack of a computational method to alleviate this problem.
\section{Conclusion \label{sec:conclusion}}
We have studied different models based on neural networks to implement a Persian reverse dictionary. All of the models take a phrase as input and determine the closest word(s) to it (called suggestions). We extracted descriptive phrases along with their corresponding words from Persian dictionaries to evaluate the models. For convenience, we refer to the corresponding words as \textit{dictionary words} in this section.

The results of the evaluation based on MOS values show that while the Bag of Words (BOW) model usually submits suggestions that are irrelevant or slightly relevant to the input phrase, the RNN model is capable of producing suggestions comparable to the dictionary words in terms of the ability to convey the concepts within the input phrase. Utilizing the RNN (LSTM) model with an additive attention mechanism also resulted in high quality suggestions being even better than the dictionary words in some cases. This phenomenon was more evident when the input phrase was extracted from the Dehkhoda dictionary.

To further improve the functionality of the model, we replaced the unidirectional LSTM with a bidirectional one (BiLSTM) since theoretically it could be less prone to gradient vanishing. However, the alteration was not followed by any clear advantage over the previous model based on the accuracy, synonym accuracy, or the MOS values. Hence, we recommend using the unidirectional LSTM along with the additive attention mechanism to bring a trade-off between the model complexity and functionality.

We have published our code on Github \footnote{\href{https://github.com/AUT-Data-Group/PREDICT-Persian-Reverse-Dictionary}{https://github.com/AUT-Data-Group/PREDICT-Persian-Reverse-Dictionary}} and our data on Kaggle \footnote{\href{https://www.kaggle.com/malekzadeharman/persian-reverse-dictionary-dataset}{https://www.kaggle.com/malekzadeharman/persian-reverse-dictionary-dataset}}. Although the proposed model successfully maps the input phrases to the appropriate words, we believe there are still ways to boost its performance. Future studies should investigate the effect of using sense embeddings instead of a unified vector for all of the senses of a word. In addition, a more complex system could predict the part-of-speech tag or semantic category of the output word based on the input.

\begin{table}[p]
	\centering
	\caption{The accuracy, synonym accuracy, and cosine loss of the models on the seen and unseen data based on the number of recognized output words}
	\resizebox{\linewidth}{!}{%
		\begin{tabular}{l|c|c|c|c|c|c|c|c|c|}
			\cline{2-10}
			\multicolumn{1}{c|}{\multirow{2}{*}{}} & \multicolumn{2}{c|}{BOW} & \multicolumn{2}{c|}{RNN} & \multicolumn{2}{c|}{LSTM+att} & \multicolumn{2}{c|}{BiLSTM+att} & \multirow{2}{*}{Number of Words} \\ \cline{2-9}
			\multicolumn{1}{c|}{}                  & training       & testing      & training       & testing      & training         & testing         & training          & testing          &                                  \\ \hlineB{3}
			\multicolumn{1}{|l|}{acc@10}           & 0.05       & 0.07        & 0.22       & 0.12        & 0.32         & 0.15           & 0.35          & 0.15            & \multirow{4}{*}{3000}            \\ \cline{1-9}
			\multicolumn{1}{|l|}{acc@100}          & 0.27       & 0.12        & 0.57       & 0.30        & 0.62         & 0.42           & 0.70          & 0.37            &                                  \\ \cline{1-9}
			\multicolumn{1}{|l|}{syn-acc@10}       & 0.30       & 0.20        & 0.42       & 0.25        & 0.47         & 0.27           & 0.5           & 0.25            &                                  \\ \cline{1-9}
			\multicolumn{1}{|l|}{syn-acc@100}      & 0.60       & 0.52        & 0.82       & 0.70        & 0.82         & 0.70           & 0.85          & 0.72            &   
			\\ \cline{1-9}
			\multicolumn{1}{|l|}{cosine loss}      & 0.54       & 0.63        & 0.51       & 0.61        & 0.48         & 0.61           & 0.49          & 0.61            &                   \\ \hlineB{4}
			\multicolumn{1}{|l|}{acc@10}           & 0.10       & 0.02        & 0.15       & 0.12        & 0.20         & 0.17           & 0.22          & 0.17            & \multirow{4}{*}{5000}            \\ \cline{1-9}
			\multicolumn{1}{|l|}{acc@100}          & 0.27       & 0.15        & 0.40       & 0.20        & 0.42         & 0.25           & 0.47          & 0.30            &                                  \\ \cline{1-9}
			\multicolumn{1}{|l|}{syn-acc@10}       & 0.37       & 0.30        & 0.35       & 0.35        & 0.47         & 0.40           & 0.47          & 0.37            &                                  \\ \cline{1-9}
			\multicolumn{1}{|l|}{syn-acc@100}      & 0.65       & 0.60        & 0.67       & 0.72        & 0.77         & 0.72           & 0.77          & 0.70            &    
			\\ \cline{1-9}
			\multicolumn{1}{|l|}{cosine loss}      & 0.55       & 0.63        & 0.54       & 0.61        & 0.52         & 0.61           & 0.52          & 0.61            &                   \\ \hlineB{4}
			\multicolumn{1}{|l|}{acc@10}           & 0.12       & 0.05        & 0.12       & 0.07        & 0.17         & 0.12           & 0.17          & 0.12            & \multirow{4}{*}{10000}           \\ \cline{1-9}
			\multicolumn{1}{|l|}{acc@100}          & 0.22       & 0.12        & 0.40       & 0.27        & 0.42         & 0.32           & 0.42          & 0.32            &                                  \\ \cline{1-9}
			\multicolumn{1}{|l|}{syn-acc@10}       & 0.40       & 0.37        & 0.30       & 0.35        & 0.37         & 0.40           & 0.40          & 0.42            &                                  \\ \cline{1-9}
			\multicolumn{1}{|l|}{syn-acc@100}      & 0.57       & 0.60        & 0.65       & 0.67        & 0.65         & 0.70           & 0.65          & 0.72            &                    
			\\ \cline{1-9}
			\multicolumn{1}{|l|}{cosine loss}      & 0.56       & 0.63        & 0.55       & 0.62        & 0.55         & 0.62           & 0.54          & 0.62            &                   \\
			\hlineB{4}
			\multicolumn{1}{|l|}{acc@10}           & 0.15       & 0.12        & 0.12       & 0.15        & 0.12         & 0.15           & 0.12          & 0.12            &
			\multirow{4}{*}{20000}           \\ \cline{1-9}
			\multicolumn{1}{|l|}{acc@100}          & 0.25       & 0.25        & 0.27       & 0.27        & 0.17         & 0.22           & 0.22          & 0.22            &                                  \\ \cline{1-9}
			\multicolumn{1}{|l|}{syn-acc@10}       & 0.42       & 0.45        & 0.30       & 0.45        & 0.35         & 0.50           & 0.37          & 0.45            &                                  \\ \cline{1-9}
			\multicolumn{1}{|l|}{syn-acc@100}      & 0.57       & 0.67        & 0.45       & 0.72        & 0.47         & 0.62           & 0.52          & 0.62            &                    
			\\ \cline{1-9}
			\multicolumn{1}{|l|}{cosine loss}      & 0.57       & 0.63        & 0.56       & 0.62        & 0.57         & 0.62           & 0.57          & 0.62            &  
			\\ \hlineB{}
		\end{tabular}
	}
	\label{tab:acc-syn}
\end{table}
\begin{table}[p]
	\centering
	\caption{The quality of the suggestions of the models vs. the quality of  original words existing in Persian dictionaries}
	\resizebox{\linewidth}{!}{%
		\begin{tabular}{l|c|c|c|c|c|c|c|c|c|c|c|c|c|c|c|c|c|}
			\cline{2-18}
			\multicolumn{1}{c|}{\multirow{2}{*}{}} & \multicolumn{4}{c|}{BOW} & \multicolumn{4}{c|}{RNN} &
			\multicolumn{4}{c|}{LSTM+att} & \multicolumn{4}{c|}{BiLSTM+att} &
			\multirow{2}{*}{\# Words} \\ \cline{2-17}
			\multicolumn{1}{c|}{}                  & $q_t$   & $q_1$ & $q_2$ & $q_3$ & $q_t$   & $q_1$ & $q_2$ & $q_3$ & $q_t$     & $q_1$  & $q_2$  & $q_3$  & $q_t$     & $q_1$  & $q_2$  & $q_3$  &                                 \\ \hlineB{4}
			\multicolumn{1}{|l|}{Amid}             & 3.2 & 1.7  & 1.6  & 1.8  & 3.1 & 2.8  & 2.5  & 2.0  & 2.8   & 2.6   & 2.3   & 2.1   & 2.5   & 2.4    & 1.9    & 2.1   & \multirow{3}{*}{3000}            \\ \cline{1-17}
			\multicolumn{1}{|l|}{Dehkhoda}         & 1.6 & 1.6  & 1.3  & 1.6  & 2.1 & 1.9  & 1.8  & 1.6  & 2.0   & 1.6   & 1.5   & 1.5   & 2.1   & 2.0    & 1.6    & 1.8   &                                  \\ \cline{1-17}
			\multicolumn{1}{|l|}{Moin}             & 2.9 & 1.4  & 1.2  & 1.2  & 2.7 & 2.0  & 2.3  & 2.2  & 2.7   & 2.3   & 2.2   & 1.9   & 3.3   & 2.5    & 2.2    & 2.1   &                                  \\ \hlineB{4}
			\multicolumn{1}{|l|}{Amid}             & 3.8 & 2.3  & 2.3  & 1.8  & 3.7 & 3.2  & 2.8  & 2.4  & 3.6   & 3.0   & 2.3   & 2.4   & 3.3   & 2.3    & 2.3    & 1.5   & \multirow{3}{*}{5000}            \\ \cline{1-17}
			\multicolumn{1}{|l|}{Dehkhoda}         & 2.4 & 1.9  & 1.7  & 1.6  & 3.3 & 2.6  & 2.4  & 2.1  & 2.0   & 1.9   & 1.8   & 1.6   & 2.3   & 2.3    & 1.9    & 1.8   &                                  \\ \cline{1-17}
			\multicolumn{1}{|l|}{Moin}             & 3.4 & 1.7  & 1.5  & 2.0  & 3.3 & 2.6  & 2.4  & 2.1  & 3.7   & 2.7   & 2.1   & 1.9   & 3.1   & 2.4    & 2.1    & 2.0   &                                  \\ \hlineB{4}
			\multicolumn{1}{|l|}{Amid}             & 3.4 & 2.4  & 2.1  & 2.2  & 3.3 & 3.1  & 2.7  & 2.5  & 3.2   & 3.1   & 2.8   & 2.7   & 3.7   & 2.6    & 2.2    & 2.6   & \multirow{3}{*}{10000}           \\ \cline{1-17}
			\multicolumn{1}{|l|}{Dehkhoda}         & 2.4 & 2.2  & 1.6  & 2.0  & 1.9 & 2.5  & 2.0  & 2.0  & 2.1   & 2.2   & 2.0   & 1.9   & 2.3   & 2.4    & 2.1    & 1.8   &                                  \\ \cline{1-17}
			\multicolumn{1}{|l|}{Moin}             & 3.3 & 2.1  & 1.9  & 1.7  & 3.2 & 2.6  & 2.4  & 1.9  & 3.1   & 2.8   & 2.3   & 2.2   & 3.2   & 3.1    & 2.4    & 2.0   &                                  \\ \hlineB{4}
			\multicolumn{1}{|l|}{Amid}             & 3.1 & 2.4  & 2.6  & 2.0  & 3.2 & 2.9  & 2.8  & 2.6  & 3.1   & 2.6   & 2.3   & 2.1   & 3.2   & 2.7    & 2.5    & 2.5 & \multirow{3}{*}{20000}           \\ \cline{1-17}
			\multicolumn{1}{|l|}{Dehkhoda}         & 1.8 & 2.2  & 1.9  & 1.6  & 1.9 & 2.2  & 1.8  & 1.7  & 1.9   & 2.2   & 2.0   & 1.7   & 2.3   & 2.3    & 1.7    & 1.6   &                                  \\ \cline{1-17}
			\multicolumn{1}{|l|}{Moin}             & 2.9 & 2.6  & 2.0  & 2.1  & 3.1 & 2.8  & 2.4  & 2.1  & 2.7   & 2.5   & 2.2   & 2.0   & 2.5   & 2.5    & 2.0    & 2.2   &                                  
			\\ \hlineB{}
		\end{tabular}
	}
	\label{tab:mos-eval}
\end{table}
\section{Acknowledgements}
We are grateful to Dr. Mohammad Bahrani (Allameh Tabataba'i University) for his consultation on corpus linguistics.
We thank Dr. Maryam Danay Tous (University of Guilan) for enriching us with her knowledge in linguistics. Also, we thank the linguistics and Persian literature students of the University of Guilan for helping us to evaluate the proposed architectures using the mean opinion score.
\bibliographystyle{plainnat}  






\end{document}